%% file: neurips_2026.tex
\documentclass{article}

\usepackage[main,final]{neurips_2026}

\usepackage[utf8]{inputenc} 
\usepackage[T1]{fontenc}    
\usepackage[table]{xcolor}  
\usepackage{hyperref}       
\definecolor{lightblueref}{rgb}{0.30,0.55,0.85}
\hypersetup{
  colorlinks=true,
  citecolor=lightblueref, 
  linkcolor=red,          
  urlcolor=lightblueref
}
\usepackage{url}            
\usepackage{booktabs}       
\usepackage{amsfonts}       
\usepackage{nicefrac}       
\usepackage{microtype}      
\usepackage{graphicx}       
\usepackage{amsmath}        
\usepackage{multirow}       

\title{Beyond Single Expert: Harmonizing Diverse Visual Priors in MLLMs for Spatial Understanding}

\author{%
  Xiao Lin \qquad\qquad Xiaohu Huang \qquad\qquad Kai Han\thanks{Corresponding author.} \\
  The University of Hong Kong \\
  \texttt{\{lllinxiao, huangxiaohu\}@connect.hku.hk, kaihanx@hku.hk}
}

\begin{document}

\maketitle

\input{chapters/abstract.tex}

\input{chapters/intro.tex}

\input{chapters/method.tex}

\input{chapters/exp.tex}

\input{chapters/related_work.tex}
\input{chapters/conclusion.tex}
\input{chapters/ack.tex}

\clearpage

{
\small
\bibliographystyle{plainnat}
\bibliography{refs}
}

\newpage
\appendix
\input{chapters/appendix.tex}

\end{document}

%% file: chapters/abstract.tex
\begin{abstract}
Multimodal Large Language Models (MLLMs) have demonstrated substantial promise in spatial understanding. Existing works typically incorporate prior knowledge extracted from a pre-trained foundation model to further enhance the spatial awareness of MLLMs. In this paper, we first reveal that when integrating diverse foundation models into MLLMs, different models provide complementary spatial priors that benefit different tasks. Motivated by this, we propose \textbf{ViPS}, a novel multi-model prior framework designed to fully unleash the potential of incorporating multiple \textbf{Vi}sual \textbf{P}riors from diverse models into MLLMs for \textbf{S}patial understanding. Specifically, ViPS introduces an Efficient Prior Proxy to generate multiple foundational priors with minimal inference overhead, and a Dynamic Prior Fusion mechanism to achieve harmonious and context-aware prior fusion and injection from the prior proxies. Extensive experiments demonstrate that ViPS successfully harmonizes diverse visual priors, establishing new state-of-the-art performance across multiple complex spatial reasoning and 3D spatial understanding benchmarks. Project page: \url{https://visual-ai.github.io/vips}
\end{abstract}

%% file: chapters/intro.tex
\begin{figure}[h]
  \centering
  \includegraphics[width=\linewidth]{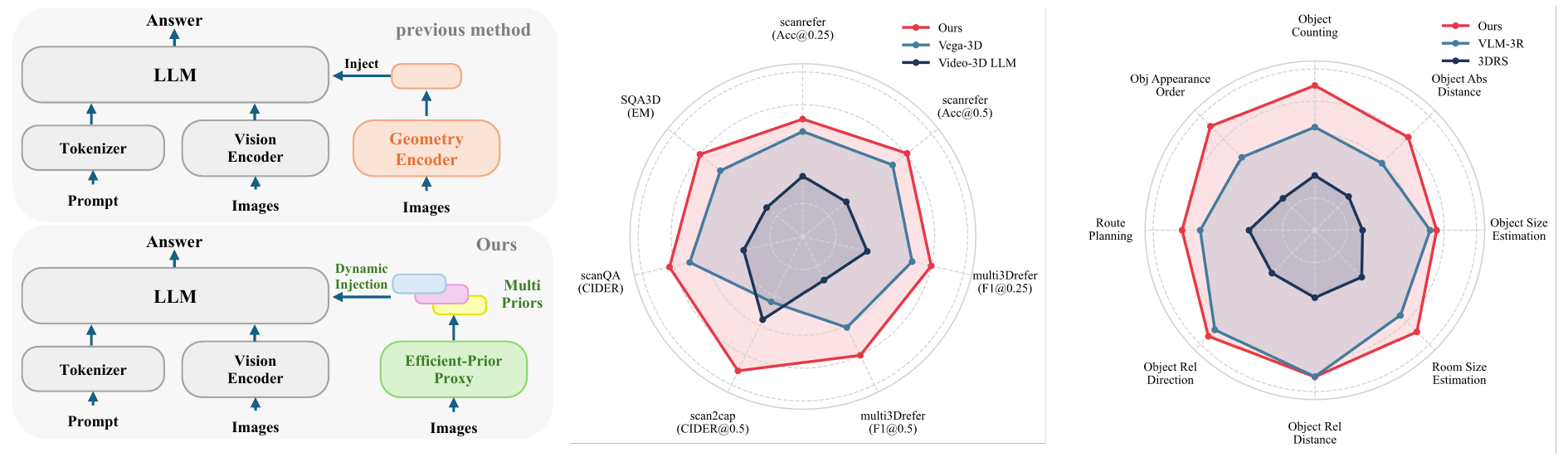}
  \caption{\textbf{Comparison of Existing Single-Expert Paradigms and Our Multi-Prior Framework.} \textbf{Top-left:} Existing paradigms typically rely on a single external encoder (e.g., VGGT) to provide visual priors for MLLMs. \textbf{Bottom-left:} In contrast, our approach integrates diverse knowledge from multiple expert models into the MLLM. \textbf{Middle and right:} Extensive evaluations demonstrate that our method achieves state-of-the-art performance across multiple benchmarks.}
  \label{fig:figure1}
\end{figure}

\section{Introduction}
Spatial understanding serves as the foundation for real-world reasoning and interaction, acting as a pivotal technology for critical applications such as robotic navigation and embodied intelligence. The rapid advancement of Multimodal Large Language Models (MLLMs)\citep{wang2024qwen2,li2024llava,chen2024internvl,hurst2024gpt,team2024gemini,bai2025qwen3} has brought substantial breakthroughs to this domain. By bridging the general reasoning capabilities of Large Language Models (LLMs) with visual or point-cloud encoders~\citep{xu2024pointllm,qi2024gpt4point}, MLLMs facilitate sophisticated spatial perception and reasoning within complex environments.

Recently, numerous studies have explored incorporating knowledge from pre-trained foundation models (e.g., VGGT~\citep{wang2025vggt}, WAN~\citep{wan2025wan}) into MLLMs~\citep{huang20253drs, wu2026generation, li2026thinking, zheng2025learning}. Typically, this paradigm involves leveraging foundational features as auxiliary inputs or aligning the latent representations of MLLMs with foundation models through distillation or feature injection. Such a paradigm has proven effective in augmenting MLLMs with domain-specific expert knowledge to boost spatial awareness. Despite their success, existing efforts are predominantly restricted to exploiting prior knowledge from a single expert, while overlooking the potential of fusing diverse knowledge from multiple foundation models. Intuitively, different foundation models capture distinct priors shaped by their varied pre-training data and objectives, making it highly desirable to aggregate their respective advantages in MLLMs. Therefore, how to synergistically harmonize these complementary strengths from distinct models remains a significant, yet unresolved challenge.

In this paper, we seek to explore two questions that form our core motivations: (i) \textit{Do different foundation models exhibit distinct specializations across various spatial understanding tasks when serving as priors for MLLMs?} (ii) \textit{How can we synergistically integrate the expert knowledge from multiple foundation models into MLLMs?} Regarding the first question, we conduct comprehensive empirical studies by integrating the priors of different foundation models~\citep{wang2025vggt, lin2025depth, liu2025trace, wan2025wan, heinrich2025radiov2} into MLLMs across a wide range of spatial understanding tasks~\citep{yang2025thinking, chen2020scanrefer, zhang2023multi3drefer, chen2021scan2cap, azuma2022scanqa, ma2022sqa3d}. Our results reveal that these models contribute differently to specific tasks, indicating a strong complementarity among diverse visual priors and underscoring the necessity of multi-model integration. Based on these findings, a straightforward solution to the second question would be a naive ensemble of multiple priors. However, such an approach suffers from several critical drawbacks: First, the computational overhead becomes prohibitive, as extracting priors from each foundation model requires an independent forward pass during inference. Second, since different visual priors inherently exhibit distinct strengths and specializations, a naive ensemble not only fails to prioritize the most relevant prior knowledge for a given task, but also risks confusing the MLLM due to the distribution disparities among different foundation models.

To address these challenges, we propose \textbf{ViPS} (illustrated in Figure~\ref{fig:figure1}), a novel multi-model prior framework designed to fully unleash the potential of multiple \textbf{Vi}sual \textbf{P}riors in MLLMs for \textbf{S}patial understanding. ViPS harmoniously integrates priors from multiple models into the MLLM by introducing an Efficient Prior Proxy and a Dynamic Prior Fusion mechanism. Specifically, the Efficient Prior Proxy is designed to generate multiple foundational priors without requiring multiple forward passes of foundation models. Instead of directly deploying all foundation models independently, we employ one base model alongside several lightweight proxies to estimate the priors of the foundation models and align the proxy outputs with the ground-truth priors during training. This design is motivated by the insight that different foundation models often share common low- and mid-level visual and geometric features. Therefore, a robust base model providing sufficient representations is enough, from which other distinct priors can be effectively distilled using lightweight proxies. Furthermore, the Dynamic Prior Fusion is proposed to achieve harmonious and context-aware prior selection from proxies. To this end, we first generate dynamic weights based on the input task, and then use these weights to aggregate the priors from different foundation models into a fused multi-expert prior, which is subsequently injected into the MLLM. Additionally, a set of zero-initialized convolutional layers is applied before aggregation to ensure that diverse priors are harmoniously fused. Extensive experiments demonstrate that ViPS achieves state-of-the-art performance across diverse spatial reasoning and spatial understanding benchmarks.

To summarize, we make the following contributions: \textit{First}, we conduct an empirical study on integrating diverse foundation models into MLLMs, revealing that different visual priors exhibit distinct specializations across various tasks. \textit{Second}, we propose \textbf{ViPS}, a novel multi-model prior framework for spatial understanding, which features an Efficient Prior Proxy to generate multi-model priors with minimal overhead, and a Dynamic Prior Fusion mechanism for harmonious and context-aware prior integration. \textit{Third}, extensive experiments demonstrate that \textbf{ViPS} achieves new state-of-the-art performance across multiple spatial reasoning and spatial understanding benchmarks.

%% file: chapters/method.tex
\begin{figure}[b]
  \centering
  \includegraphics[width=\linewidth]{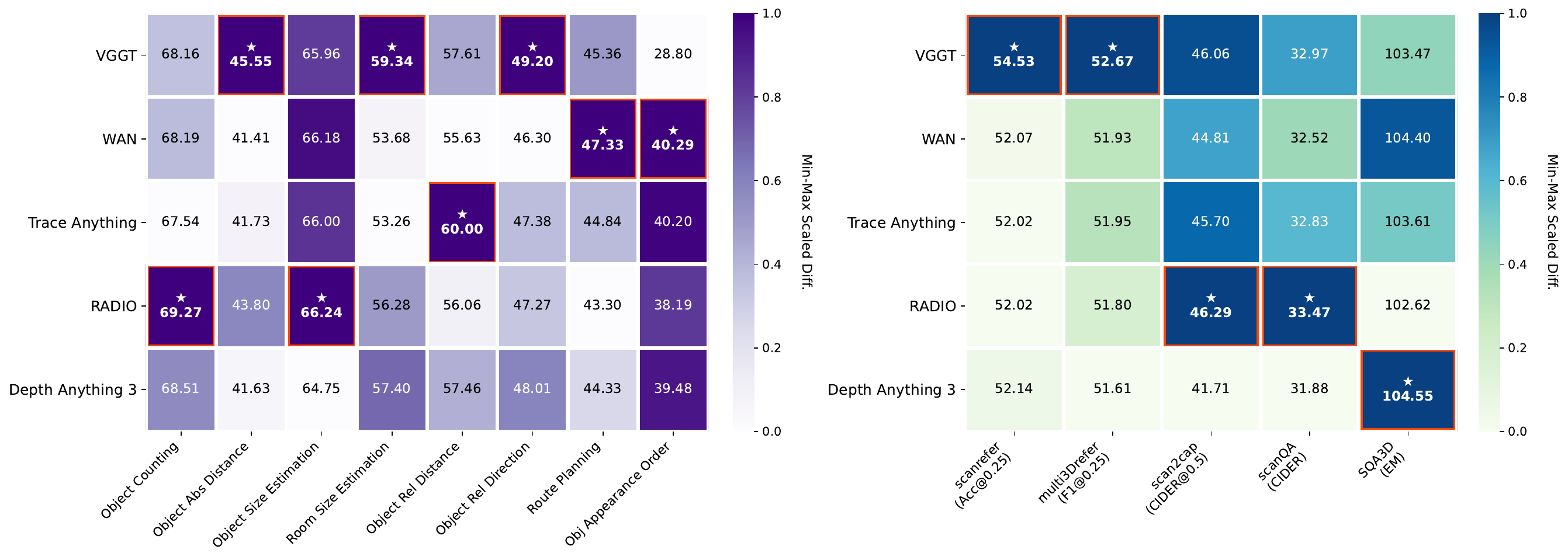}
  \caption{\textbf{Relative Performance of Diverse Foundation Models.} We evaluate the performance of various foundation models when serving as priors for MLLMs across a wide range of spatial understanding tasks. Notably, no single model dominates all metrics, thoroughly motivating the need for the multi-model prior integration.}
  \label{fig:empirical_study}
\end{figure}

\section{Method}

\subsection{Empirical Study on Diverse Model Priors}
\label{sec:motivation}
\paragraph{Preliminary.}
Our work focuses on spatial understanding based on MLLMs. Formally, given a visual input comprising a sequence of frames (e.g., a video clip or multi-view images of a scene) $\mathcal{V} = \{v_1, v_2, \dots, v_T\} \in \mathbb{R}^{T \times H \times W \times 3}$, alongside a corresponding textual task instruction $\mathcal{X}_{text} = \{x_1, x_2, \dots, x_N\}$, the objective of the MLLM is to generate a coherent and accurate textual response $\mathcal{Y} = \{y_1, y_2, \dots, y_M\}$ that adequately answers the query by reasoning over the holistic spatial structures of the scene. In standard video-based MLLM architectures (e.g., LLaVA~\citep{li2024llava}, Qwen2-VL~\citep{wang2024qwen2}), the visual sequence $\mathcal{V}$ is first processed by a generic vision encoder $\Phi_{vis}$, producing visual embeddings $F_{vis} = \Phi_{vis}(\mathcal{V})$. A modality projector $\mathcal{P}$ then aligns these embeddings into the text semantic space. The LLM, denoted as $f_{LLM}$, takes the concatenated tokens of the projected visual features and the text instruction to auto-regressively predict the response:
\begin{equation}
    P(\mathcal{Y} \mid \mathcal{V}, \mathcal{X}_{text}) = \prod_{i=1}^{M} P \left( y_i \mid \mathcal{P}(F_{vis}), \mathcal{X}_{text}, y_{<i}; \theta \right),
\end{equation}
where $\theta$ encapsulates the learnable parameters of the system. To bridge the gap in 3D spatial awareness, recent approaches explicitly introduce external foundation model priors. Given a pre-trained visual expert $\mathcal{E}$ (e.g., encoders from foundation models like VGGT), it extracts domain-specific prior features $F_{prior} = \mathcal{E}(\mathcal{V})$. The LLM leverages this supplementary knowledge to enhance generation, yielding $P(\mathcal{Y} \mid F_{vis}, F_{prior}, \mathcal{X}_{text})$. While integrating a single prior model provides distinct utility, relying exclusively on an individual $\mathcal{E}$ inevitably limits the representation capacity, forming the motivation for our extensive empirical investigations.

\paragraph{Key Finding and Motivation.}
As highlighted in the introduction, existing efforts are restricted to exploiting the priors of a single model, overlooking the potential of fusing diverse knowledge from multiple foundation models. Fundamentally, different foundation models encapsulate distinct representations and spatial semantics, thereby endowing them with disparate advantages when serving as priors for MLLMs. To demonstrate this, we select a diverse set of foundation models for evaluation, including VGGT~\citep{wang2025vggt}, DepthAnything3~\citep{lin2025depth}, TraceAnything~\citep{liu2025trace}, Wan2.1~\citep{wan2025wan}, and RADIO~\citep{heinrich2025radiov2}. Their feature distributions and activation patterns are analyzed in Appendix~\ref{sec:prior_analysis}.
Subsequently, we explore their relative advantages across a wide range of spatial understanding tasks when serving as priors for MLLMs (See \ref{sec:motivation_details} for more details). The results are shown in Figure~\ref{fig:empirical_study}. From a column-wise perspective, the best-performing model varies across different sub-tasks. From a row-wise perspective, each specific model achieves its optimal performance on disparate sub-tasks. These results demonstrate that no single foundation model achieves universal dominance; instead, different models exhibit distinct strengths and specializations. The highly complementary nature of these diverse visual priors directly motivates our core objective: to efficiently extract and adaptively harmonize distinct prior knowledge from multiple foundation models for comprehensive spatial reasoning.

\subsection{Proposed Framework: ViPS}
\begin{figure}[t]
  \centering
  \includegraphics[width=\linewidth]{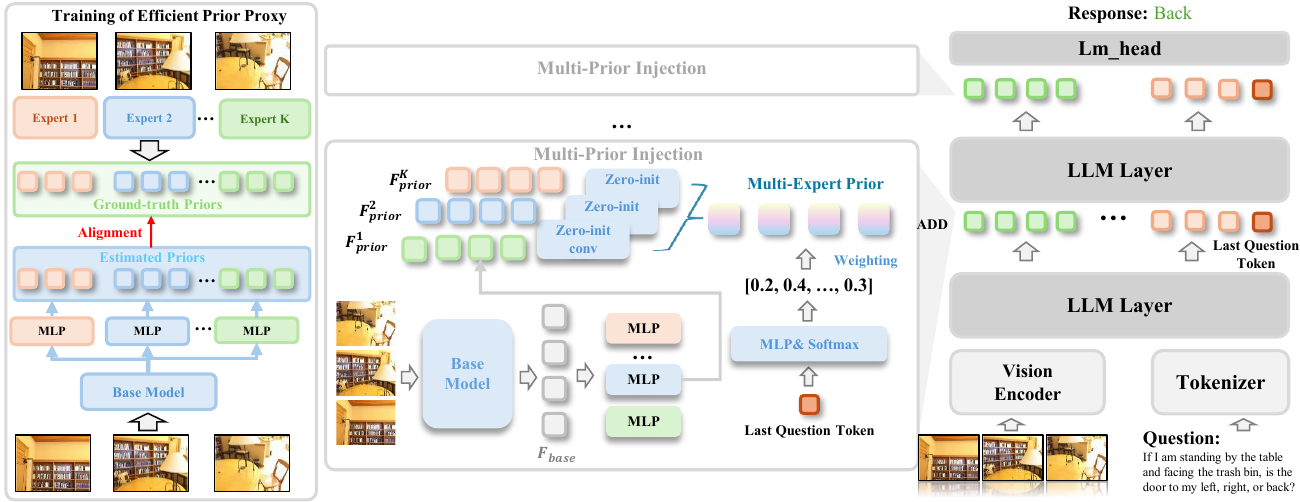}
  \caption{\textbf{Overview of the Proposed ViPS Framework.} The framework integrates distinct prior knowledge from multiple foundation models via the Efficient Prior Proxy and coordinates them using Dynamic Prior Fusion for comprehensive spatial reasoning.}
  \label{fig:method}
\end{figure}

Building upon the above insights, we introduce \textbf{ViPS} (\textbf{Vi}sual \textbf{P}riors for \textbf{S}patial understanding), a novel framework designed to fully unleash the potential of leveraging diverse prior models in MLLMs. The overall framework is illustrated in Figure~\ref{fig:method}. ViPS first processes the input visual sequence $\mathcal{V}$ through a standard vision encoder $\Phi_{vis}$~\citep{zhai2023sigmoid}, producing the base visual embeddings $F_{vis}$. Instead of relying on a single expert or naively ensembling multiple heavy models, ViPS seamlessly integrates comprehensive prior knowledge through two pivotal mechanisms: \textbf{Efficient Prior Proxy} and \textbf{Dynamic Prior Fusion}. 

Specifically, the Efficient Prior Proxy utilizes a base foundation model alongside lightweight MLPs to efficiently estimate the prior features of various external models, generating a set of diverse prior representations $\{F^1_{prior}, F^2_{prior}, \dots, F^K_{prior}\}$. Subsequently, these representations are fed into the Dynamic Prior Fusion part, which first applies independent, zero-initialized convolutional layers~\citep{zhang2023adding} to each prior branch and then employs a context-aware weighting mechanism, guided by the final token of the input instruction $\mathcal{X}_{text}$, to dynamically compute fusion weights. Finally, the harmoniously fused multi-expert prior $\hat{F}_{prior}$ is incorporated into the MLLM, empowering the LLM to leverage the most relevant visual knowledge tailored to the specific spatial reasoning task. We detail the formulations of these two components below.

\paragraph{Efficient Prior Proxy.}
Our framework aims to effectively integrate highly complementary priors from various foundation models for comprehensive spatial understanding. Directly extracting visual features from a set of $K$ distinct foundation models, denoted as $\{\mathcal{E}_1, \mathcal{E}_2, \dots, \mathcal{E}_K\}$, conventionally requires $K$ independent forward passes, i.e., computing $\mathcal{E}_k(\mathcal{V})$ for $k \in \{1, \dots, K\}$. This paradigm introduces a severe computational bottleneck, as the inference latency and memory footprint scale linearly with the number of integrated priors. To decouple the computational overhead from the integration of multiple models while effectively preserving the rich diversity of visual prior knowledge, we propose the efficient prior proxy mechanism. Specifically, we use a single robust vision encoder as the base model $\mathcal{E}_{base}$ to extract a unified foundational representation:
\begin{equation}
    F_{base} = \mathcal{E}_{base}(\mathcal{V}) \in \mathbb{R}^{S \times D_{base}},
\end{equation}
where $S$ and $D_{base}$ denote the sequence length and channel dimension of the visual tokens, respectively. In our experiments, the base model is directly initialized with the encoder of one of the foundation models (we present experiments using encoders from different foundation models as the base model in the appendix). 

To approximate the distinct knowledge of the $K$ targeted foundation models, we instantiate a set of $K$ lightweight proxy networks. Each proxy $\phi_k(\cdot)$ is implemented as a simple Multi-Layer Perceptron (MLP) designed to extract specific prior semantics (e.g., depth cues, geometric boundaries, or detailed semantics) from the shared foundational feature $F_{base}$. The representation for the $k$-th prior is uniformly formulated as:
\begin{equation}
    F^k_{prior} = \phi_k(F_{base}) \in \mathbb{R}^{S \times D_{prior}},
\end{equation}
where $D_{prior}^{k}$ is the feature dimension for the $k$-th prior feature. 

To guarantee the fidelity of these estimated prior representations, we explicitly supervise the output of prior proxies during the training phase. Specifically, the features extracted by the $K$ distinct foundation models are utilized as the ground-truth targets, denoted as $F^k_{gt} = \mathcal{E}_k(\mathcal{V})$. We apply an $L_2$ loss to enforce strict alignment between the proxy outputs $F^k_{prior}$ and their corresponding ground-truth priors $F^k_{gt}$:
\begin{equation}
    \mathcal{L}_{alignment} = \sum_{k=1}^K \big\| F^k_{prior} - F^k_{gt} \big\|_2^2.
\end{equation}

The viability of this proxy-based estimation stems from the intuition that different foundation models inherently share substantial low- and mid-level visual semantics; thus, a robust base representation contains sufficient foundational knowledge from which distinct, high-level prior semantics can be efficiently derived via shallow proxies. Given that the computational cost of a shallow MLP $\phi_k$ is negligible compared to a full foundation model, our framework can scale to an arbitrary number of diverse priors without linearly increasing inference overhead, while ensuring that the estimated visual priors remain accurate through the alignment loss.

\paragraph{Dynamic Prior Fusion.}
With the diverse prior representations $\{F^1_{prior}, \dots, F^K_{prior}\}$ efficiently established, the remaining challenge is seamlessly integrating them into the MLLM. Directly adding these distinct and heterogeneous priors fails to prioritize the most effective priors for the specific input task, and can overwhelm the language model, consequently degrading overall performance. To achieve harmonious and context-aware integration, we propose the Dynamic Prior Fusion mechanism. 

Accordingly, we design a dynamic weighting mechanism to selectively focus on the precise knowledge required by the given task. Specifically, we extract the representation of the final token of the text query, denoted as $x_{last}^{text}$, which naturally encapsulates the aggregated semantics of the entire instruction due to the causal mechanism of the LLM. This context vector is passed through an MLP to compute a set of $K$-dimensional logits $z \in \mathbb{R}^K$:
\begin{equation}
    z = \text{MLP}_{weight}(x_{last}^{text}).
\end{equation}
A Softmax activation is subsequently applied to $z$ to obtain the normalized fusion weights $w \in [0, 1]^K$. These weights will be utilized to dynamically combine the diverse priors. However, our experiments show that directly doing so fails to yield progressive performance improvements. We attribute this phenomenon to the disparate prior distributions originating from different source models, which tend to confuse the MLLM during the early stages (see Appendix~\ref{sec:prior_analysis} and Section~\ref{sec:ablation}). To overcome this issue and enable progressive prior injection, we apply an independent zero-initialized convolutional layer, denoted as $\text{ZeroConv}_k(\cdot)$, to each proxy branch prior to the weighted fusion. This strategy ensures that the outputs of the convolution are initially zero, preserving the original MLLM representations and preventing the prior perturbations from disrupting early training, while gradually scaling up as the MLLM progressively learns from the diverse prior knowledge. The processed prior $\tilde{F}^k_{prior}$ is computed as:
\begin{equation}
    \tilde{F}^k_{prior} = \text{ZeroConv}_k(F^k_{prior}).
\end{equation}

Finally, the context-aware weights are utilized to linearly combine the aligned priors, yielding the overall harmonized multi-expert prior $\hat{F}_{prior}$:
\begin{equation}
    \label{eq:prior_fusion}
    \hat{F}_{prior} = \sum_{k=1}^K w_k \cdot \tilde{F}^k_{prior}.
\end{equation}
This fused representation $\hat{F}_{prior}$, which dynamically captures the most pertinent visual prior requested by the query, is directly added element-wise to the corresponding projected visual embeddings $\mathcal{P}(F_{vis})$ (i.e., the image tokens) of the MLLM. In this way, Dynamic Prior Fusion seamlessly integrates diverse priors into the MLLM, adaptively tailoring knowledge selection to the input task while leveraging zero-initialization to ensure progressive learning without disrupting early training. 

It is worth noting that, for simplicity of presentation, the above formulation describes the prior injection process at a single layer. In practice, we uniformly apply this Dynamic Prior Fusion mechanism across five layers of the MLLM. This multi-layer injection ensures that the diverse prior knowledge is deeply integrated, progressively guiding the spatial reasoning of the model.

%% file: chapters/exp.tex
\section{Experiments}
\subsection{Dataset and Evaluation Metric}
We evaluate our method across a total of six datasets: VSI-Bench~\citep{yang2025thinking} for spatial reasoning tasks, and five ScanNet-series benchmarks for 3D spatial understanding (ScanRefer~\citep{chen2020scanrefer}, Multi3DRefer~\citep{zhang2023multi3drefer}, Scan2Cap~\citep{chen2021scan2cap}, ScanQA~\citep{azuma2022scanqa}, and SQA3D~\citep{ma2022sqa3d}). Specifically, VSI-Bench comprises eight fine-grained spatial reasoning tasks: object counting, absolute distance estimation, object size estimation, room size estimation, relative distance, relative direction, route planning, and object appearance order. Following the standard protocol, we report the accuracy (\%) across all its sub-tasks. The five ScanNet-series benchmarks, derived from the ScanNet corpus, span three primary tasks: (i) \textit{3D visual grounding}, where ScanRefer targets free-form object localization and Multi3DRefer extends this to multi-target and zero-target ambiguity; (ii) \textit{dense captioning}, using Scan2Cap to generate natural language descriptions for localized objects; and (iii) \textit{embodied question answering}, employing ScanQA for geometry-grounded open-ended questions and SQA3D for complex situated reasoning from a specific agent perspective. Following standard protocols, we report Acc@0.25/0.5 for ScanRefer, F1@0.25/0.5 for Multi3DRefer, CIDEr@0.5 for Scan2Cap, CIDEr and Exact Match (EM) for ScanQA, and Exact Match for SQA3D. More details can be found in the Appendix.

\subsection{Implementation Detail}
We use both Qwen2-VL-7B~\citep{wang2024qwen2} and Qwen3-VL-8B~\citep{bai2025qwen3} as base MLLMs on VSI-Bench; for all other experiments, we adopt Qwen2-VL-7B as the default model. To comprehensively capture spatial knowledge, we integrate $K=5$ distinct foundation models to extract complementary visual priors. Specifically, we employ VGGT~\citep{wang2025vggt} for general visual-geometric feature extraction, which also serves as the base foundation model in the Efficient Prior Proxy. We further incorporate DepthAnything3~\citep{lin2025depth} for precise monocular depth cues, TraceAnything~\citep{liu2025trace} for object-centric motion and relationship tracking, Wan2.1~\citep{wan2025wan} for rich spatio-temporal dynamics, and RADIO~\citep{heinrich2025radiov2} for robust generic visual representations. These diverse priors collectively provide a holistic understanding of 3D scenes. We uniformly sample 32 frames from each video as the input to the vision encoder. During training, we first apply the alignment loss to train the Efficient Prior Proxy, and then freeze the proxies while fine-tuning the MLLM. We freeze the vision encoder and apply Low-Rank Adaptation (LoRA)~\citep{hu2022lora} to the LLM backbone during MLLM fine-tuning. We optimize the model using the Adam optimizer~\citep{kingma2014adam} with a batch size of 16 and a maximum learning rate of $1 \times 10^{-5}$. The model is trained for 1 epoch on each dataset. More details can be found in the Appendix. 

\subsection{Comparison with State-of-the-Art Methods}
We evaluate ViPS against leading MLLMs and specialized 3D spatial understanding models. Table~\ref{tab:vsibench_results} and Table~\ref{tab:scannet_results} summarize our results on VSI-Bench and the ScanNet-series benchmarks respectively.

\textbf{Spatial Reasoning on VSI-Bench.} As shown in Table~\ref{tab:vsibench_results}, ViPS achieves a leading average score of $63.8\%$, surpassing the previous best spatial-enhanced model VLM-3R~\citep{fan2025vlm} ($57.2\%$). While trailing VG-LLM-8B~\citep{zheng2025learning} in Appearance Order, ViPS dominates most other categories, proving the advantage of dynamically injecting multiple priors over relying solely on a single foundation model like VLM-3R or 3DRS~\citep{huang20253drs}.

\input{tables/table1}

\textbf{3D spatial understanding on ScanNet-series.} Table~\ref{tab:scannet_results} shows our results across grounding, captioning, and QA tasks. Compared to generalist 3D-LLMs like Vega-3D~\citep{wu2026generation} and 3DRS~\citep{huang20253drs}, ViPS delivers highly competitive performance. It yields top scores on visual grounding (ScanRefer, $64.6\%$ Acc@0.25) and QA (ScanQA, $107.9$ CIDEr), outperforming Vega-3D. Although ViPS slightly trails 3DRS on Scan2Cap, these results confirm that our unified proxy injection effectively harnesses multiple foundation priors for complex 3D spatial understanding tasks.

\input{tables/table2}

\subsection{Ablation Study}
\label{sec:ablation}

\textbf{Effectiveness of Individual Visual Priors.} Table~\ref{tab:ablation_priors} evaluates the impact of individual foundation models compared to a baseline MLLM without prior injection. While each single prior yields distinct performance gains, our full ViPS framework integrating all five priors achieves the highest scores across all metrics. This confirms both the effectiveness of individual priors and the strong complementarity of harmonizing heterogeneous visual knowledge for 3D spatial understanding.

\input{tables/table3}

\textbf{Effectiveness of Dynamic Prior Injection.} Table~\ref{tab:ablation_injection} compares the full ViPS model against two injection variants. First, replacing zero-initialized convolutions with standard random initialization (\textit{w/o Zero-init}) causes a severe performance drop. This confirms our motivation that zero-init is essential to prevent the diverse, unaligned prior distributions from confusing the MLLM during early training, preserving the original latent space and enabling progressive prior injection. Second, substituting the dynamic proxy fusion with a straightforward feature sum (\textit{Vanilla Addition}) degrades performance. This demonstrates that our dynamic mechanism, which adapts to the input query, achieves more effective prior injection than static feature aggregation.

\input{tables/table4}
\textbf{Effectiveness of Efficient Prior Proxy.} To demonstrate the efficiency and accuracy of our Efficient Prior Proxy, we compare our method against an upper-bound setting (\textit{w/ GT Priors}) where ground-truth features from all five foundation models are used to replace priors from the Efficient Prior Proxy. As shown in Table~\ref{tab:ablation_epp}, our method employs multiple lightweight proxies to replace full foundation models, significantly reducing parameter overhead and inference latency ($1\times$ vs. $\sim5\times$) with only a marginal drop in overall performance. We also report the estimation error (0.252 in cosine distance) between the estimated priors and the ground-truth priors during inference. These experimental results consistently demonstrate the effectiveness of our proposed approach. Furthermore, we conduct an ablation study by removing the alignment loss ($\mathcal{L}_{alignment}$) while retaining all other prior proxy structures (in this setting, the whole framework, including the MLLM and Efficient Prior Proxy, is trained end-to-end). Surprisingly, although the performance is inferior to the setting with the alignment loss, it still achieves considerable improvements compared to the baseline. We attribute this to the fact that even without the alignment loss, different prior proxies can still serve as projectors onto distinct sub-feature spaces, enabling the MLLM to learn diverse knowledge from the base model, akin to the multi-head mechanism in attention networks.

\input{tables/table5}

\subsection{Analysis of Dynamic Prior Weights}
\label{sec:weight_analysis}
\begin{figure}[h]
  \centering
  \includegraphics[width=\linewidth]{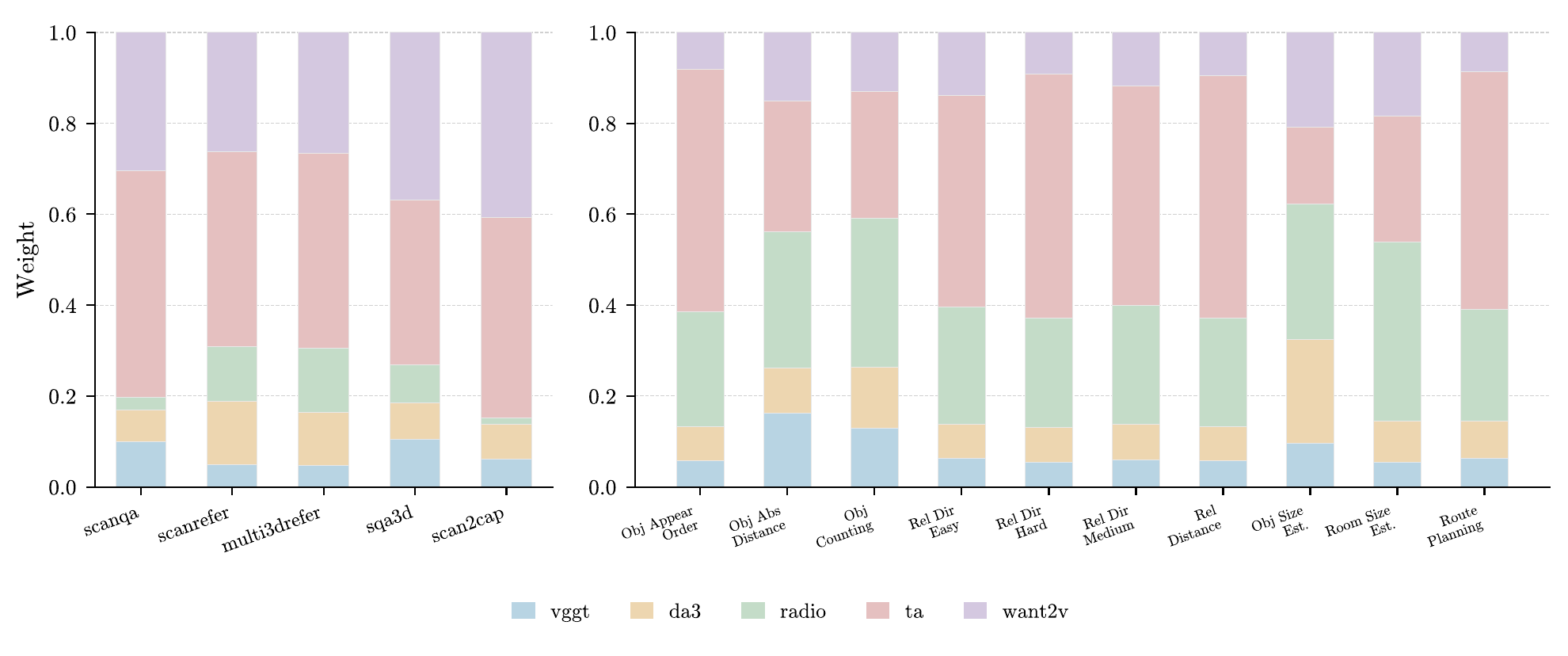}
  \caption{\textbf{Distribution of Dynamic Prior Weights.} This figure illustrates the distribution of the learned fusion weights ($w_k$ in Equation~\ref{eq:prior_fusion}) assigned to different foundation models. \textbf{Left:} The weight distribution on the test sets of ScanQA, ScanRefer, Multi3DRefer, SQA3D, and Scan2Cap. \textbf{Right:} The weight distribution across different question types in VSI-Bench.}
  \label{fig:weight_analysis}
\end{figure}

To further understand the behavior of our Dynamic Prior Fusion mechanism, we visualize the learned fusion weights ($w_k$ in Equation~\ref{eq:prior_fusion}) across different spatial understanding tasks. As illustrated in Figure~\ref{fig:weight_analysis}, the weight distribution adapts dynamically depending on different downstream tasks (e.g., ScanQA, ScanRefer, Multi3DRefer, SQA3D, and Scan2Cap), as well as across distinct question types within the same dataset (e.g., VSI-Bench). This observation aligns with our motivation that different foundation models exhibit distinct specializations, which shows that our framework successfully achieves context-aware prior selection by dynamically adjusting the fusion weights based on the specific scenario rather than statically relying on a single expert.

%% file: tables/table1.tex
\begin{table}[!t]
    \centering
    \caption{\textbf{Performance Comparison on VSI-Bench.} ViPS achieves a leading average score across the eight spatial reasoning sub-tasks, surpassing prior spatial-enhanced MLLMs.}
    \label{tab:vsibench_results}
    \footnotesize
    \setlength{\tabcolsep}{1.5pt}
    \begin{tabular*}{\textwidth}{@{\extracolsep{\fill}}l*{9}{c}@{}}
    \toprule
    & & \multicolumn{4}{c}{\cellcolor{red!5}Numerical Answer} & \multicolumn{4}{c}{\cellcolor{yellow!10}Multiple-Choice Answer} \\
    \cmidrule(lr){3-6} \cmidrule(lr){7-10}
    Model & Avg. & Count & Abs. Dist. & Obj. Size & Room Size & Rel. Dist. & Rel. Dir. & Route & Order \\
    \midrule
    \multicolumn{10}{l}{\cellcolor{gray!15}\textit{Proprietary Models (API)}} \\
    GPT-4o~\citep{hurst2024gpt} & 34.0 & 46.2 & 5.3 & 43.8 & 38.2 & 37.0 & 41.3 & 31.5 & 28.5 \\
    Gemini-1.5-Pro~\citep{team2024gemini} & 45.4 & 56.2 & 30.9 & 64.1 & 43.6 & 51.3 & 46.3 & 36.0 & 34.6 \\
    \midrule
    \multicolumn{10}{l}{\cellcolor{gray!15}\textit{Open-source Models}} \\
    LongVA-7B~\citep{zhang2024long} & 29.2 & 38.0 & 16.6 & 38.9 & 22.2 & 33.1 & 43.3 & 25.4 & 15.7 \\
    LongVILA-8B~\citep{chen2024longvila} & 21.6 & 29.1 & 9.1 & 16.7 & 0.0 & 29.6 & 30.7 & 32.5 & 25.5 \\
    InternVL2-8B~\citep{chen2024internvl} & 34.6 & 23.1 & 28.7 & 48.2 & 39.8 & 36.7 & 30.7 & 29.9 & 39.6 \\
    InternVL2-40B~\citep{chen2024internvl} & 36.0 & 34.9 & 26.9 & 46.5 & 31.8 & 42.1 & 32.2 & 34.0 & 39.6 \\
    VILA-1.5-40B~\citep{liu2025nvila} & 31.2 & 22.4 & 24.8 & 48.7 & 22.7 & 40.5 & 25.7 & 31.5 & 32.9 \\
    LLaVA-OneVision-7B~\citep{li2024llava} & 32.4 & 47.7 & 20.2 & 47.4 & 12.3 & 42.5 & 35.2 & 29.4 & 24.4 \\
    LLaVA-OneVision-72B~\citep{li2024llava} & 40.2 & 43.5 & 23.9 & 57.6 & 37.5 & 42.5 & 39.9 & 32.5 & 44.6 \\
    LLaVA-NeXT-Video-7B~\citep{liu2024llava} & 35.6 & 48.5 & 14.0 & 47.8 & 24.2 & 43.5 & 42.4 & 34.0 & 30.6 \\
    LLaVA-NeXT-Video-72B~\citep{liu2024llava} & 40.9 & 48.9 & 22.8 & 57.4 & 35.3 & 42.4 & 36.7 & 35.0 & 48.6 \\
    \midrule
    \multicolumn{10}{l}{\cellcolor{gray!15}\textit{Spatial-Enhanced Models}} \\
    
    vsGRPO-V-7B~\citep{liao2025improved} & 40.7 & 59.9 & 29.6 & 50.8 & 48.3 & 35.4 & 35.6 & 34.0 & 31.5 \\
    SPAR-8B~\citep{zhang2025flatland} & 41.1 & - & - & - & - & - & - & - & - \\
    SpaceR-7B~\citep{ouyang2025spacer} & 45.6 & - & - & - & - & - & - & - & - \\
    VG-LLM-4B~\citep{zheng2025learning} & 45.9 & 65.6 & 37.4 & 54.8 & 60.2 & 42.3 & 46.3 & 33.0 & 25.9 \\
    VG-LLM-8B~\citep{zheng2025learning} & 50.1 & 67.2 & 38.0 & 59.3 & 63.2 & 47.0 & 43.9 & 33.0 & 49.4 \\
    3DRS-7B~\citep{huang20253drs} & 45.9 & 68.7 & 34.8 & 53.6 & 56.6 & 40.9 & 43.2 & 30.4 & 39.2 \\
    Vega-3D~\citep{wu2026generation} & 50.5 & 69.7 & 35.9 & 58.0 & 60.8 & 45.1 & 43.1 & 30.9 & \textbf{60.5} \\
    VLM-3R~\citep{fan2025vlm} & 57.2 & 70.2 & 49.4 & 69.2 & 67.1 & 65.4 & 80.5 & 45.4 & 40.1 \\
    \midrule
    \textbf{ViPS} (qwen2-vl-7b) & \textbf{63.8} & 71.5 & \textbf{60.8} & 70.7 & \textbf{71.6} & \textbf{65.5} & \textbf{84.7} & \textbf{51.0} & 34.3 \\
    \textbf{ViPS} (qwen3-vl-8b) & \textbf{63.8} & \textbf{82.6} & 50.5 & \textbf{81.4} & 64.1 & 62.8 & 64.1 & 46.2 & \textbf{58.3} \\
    \bottomrule
    \end{tabular*}
\end{table}

%% file: tables/table2.tex
\begin{table}[!t]
\centering
\caption{\textbf{Performance Comparison on ScanNet-Series Benchmarks.} ViPS delivers competitive performance across 3D visual grounding (ScanRefer, Multi3DRefer), dense captioning (Scan2Cap), and embodied question answering (ScanQA, SQA3D).}
\label{tab:scannet_results}
\footnotesize
\setlength{\tabcolsep}{1.5pt}
\begin{tabular*}{\textwidth}{@{\extracolsep{\fill}}l*{8}{c}@{}}
\toprule
& \multicolumn{2}{c}{ScanRefer} & \multicolumn{2}{c}{Multi3DRefer} & \multicolumn{1}{c}{Scan2Cap} & \multicolumn{2}{c}{ScanQA} & \multicolumn{1}{c}{SQA3D} \\
\cmidrule(lr){2-3} \cmidrule(lr){4-5} \cmidrule(lr){6-6} \cmidrule(lr){7-8} \cmidrule(lr){9-9}
Method & Acc@0.25 & Acc@0.5 & F1@0.25 & F1@0.5 & C@0.5 & CIDEr & EM & EM \\
\midrule
\multicolumn{9}{l}{\cellcolor{gray!15}\textit{Specialists}} \\
ScanRefer~\citep{chen2020scanrefer} & 37.3 & 24.3 & -- & -- & -- & -- & -- & -- \\
M3DRef-CLIP~\citep{zhang2023multi3drefer} & 51.9 & 44.7 & 42.8 & -- & 38.4 & -- & -- & -- \\
Scan2Cap~\citep{chen2021scan2cap} & -- & -- & -- & -- & 35.2 & -- & -- & -- \\
ScanQA~\citep{azuma2022scanqa} & -- & -- & -- & -- & -- & 64.9 & 21.1 & 47.2 \\
3D-VisTA~\citep{zhu20233d} & 50.6 & 45.8 & -- & -- & 66.9 & 69.6 & 22.4 & 48.5 \\
\midrule
\multicolumn{9}{l}{\cellcolor{gray!15}\textit{Generalists}} \\
3D-LLM (BLIP2-flant5)~\citep{hong20233d} & 30.3 & -- & -- & -- & -- & 69.4 & 20.5 & -- \\
Chat-3D v2~\citep{huang2023chat} & 42.5 & 38.4 & 45.1 & 41.6 & 63.9 & 87.6 & -- & 54.7 \\
SceneLLM~\citep{fu2024scene} & -- & -- & -- & -- & -- & 80.0 & 27.2 & 53.6 \\
Grounded 3D-LLM~\citep{chen2024grounded} & 47.9 & 44.1 & 45.2 & 40.6 & 70.6 & 72.7 & -- & -- \\
PQ3D~\citep{zhu2024unifying} & 57.0 & 51.2 & -- & 50.1 & 80.3 & -- & -- & 47.1 \\
ChatScene~\citep{huang2023chat} & 55.5 & 50.2 & 57.1 & 52.4 & 77.1 & 87.7 & 21.6 & 54.6 \\
LLaVA-3D~\citep{zhu2024llava} & 54.1 & 42.4 & -- & -- & 79.2 & 91.7 & 27.0 & 55.6 \\
Inst3D-LLM~\citep{yu2025inst3d} & 57.8 & 51.6 & 58.3 & 53.5 & 79.7 & 88.6 & 24.6 & -- \\
3D-LLaVA~\citep{deng20253d} & 51.2 & 40.6 & -- & -- & 78.8 & 92.6 & -- & 54.5 \\
Video-3D LLM~\citep{zheng2025video} & 58.1 & 51.7 & 58.0 & 52.7 & 83.8 & 102.1 & 30.1 & 58.6 \\
3DRS~\citep{huang20253drs} & 62.9 & 56.1 & 60.4 & 54.9 & \textbf{86.1} & 104.8 & 30.3 & 60.6 \\
Vega-3D~\citep{wu2026generation} & 63.2 & 56.2 & 60.8 & 55.1 & 83.2 & 106.3 & 30.4 & 61.3 \\
\midrule
\textbf{ViPS} & \textbf{64.6} & \textbf{57.6} & \textbf{62.0} & \textbf{56.5} & 85.5 & \textbf{107.9} & \textbf{31.6} & \textbf{62.5} \\
\bottomrule
\end{tabular*}
\end{table}

%% file: tables/table3.tex
\begin{table*}[t]
    \centering
    \small
    \caption{\textbf{Ablation on Individual Visual Priors.} Each foundation-model prior yields distinct gains over the no-prior baseline, and combining all five priors achieves the best results across the ScanNet-series benchmarks.}
    \label{tab:ablation_priors}
    \setlength{\tabcolsep}{3.5pt}
    \begin{tabular}{l c c c c c c c c}
    \toprule
    \multirow{2}{*}{\textbf{Method}} & \multicolumn{2}{c}{ScanRefer} & \multicolumn{2}{c}{Multi3DRefer} & Scan2Cap & \multicolumn{2}{c}{ScanQA} & SQA3D \\
    \cmidrule(lr){2-3} \cmidrule(lr){4-5} \cmidrule(lr){6-6} \cmidrule(lr){7-8} \cmidrule(lr){9-9}
     & Acc@0.25 & Acc@0.5 & F1@0.25 & F1@0.5 & C@0.5 & C & EM & EM \\
    \midrule
    Baseline & 62.1 & 54.6 & 59.6 & 54.4 & 81.4 & 104.6 & 30.5 & 60.8 \\
    \midrule
    + RADIO & 62.9 & 56.0 & 61.2 & 55.5 & 82.6 & 105.1 & 30.7 & 61.1 \\
    + DepthAnything3 & 62.9 & 56.2 & 61.0 & 55.7 & 81.8 & 106.2 & 30.9 & 61.2 \\
    + TraceAnything & 62.8 & 55.8 & 60.6 & 55.2 & 82.7 & 106.7 & 30.8 & 61.4 \\
    + VGGT & 63.2 & 56.3 & 61.1 & 55.8 & 81.9 & 106.6 & 31.1 & 61.6 \\
    + Wan2.1 & 62.7 & 55.8 & 61.1 & 55.5 & 81.3 & 105.5 & 30.6 & 61.2 \\
    \midrule
    \textbf{ViPS (Full)} & \textbf{64.6} & \textbf{57.6} & \textbf{62.0} & \textbf{56.5} & \textbf{85.5} & \textbf{107.9} & \textbf{31.6} & \textbf{62.5} \\
    \bottomrule
    \end{tabular}
\end{table*}

%% file: tables/table4.tex
\begin{table*}[t]
    \centering
    \small
    \caption{\textbf{Ablation on the Dynamic Prior Injection.} Both the zero-initialized convolution and the context-aware dynamic fusion are essential for harmonizing diverse priors without disrupting early training.}
    \label{tab:ablation_injection}
    \setlength{\tabcolsep}{4.5pt}
    \begin{tabular}{l c c c c c c c c}
    \toprule
    \multirow{2}{*}{\textbf{Method}} & \multicolumn{2}{c}{ScanRefer} & \multicolumn{2}{c}{Multi3DRefer} & Scan2Cap & \multicolumn{2}{c}{ScanQA} & SQA3D \\
    \cmidrule(lr){2-3} \cmidrule(lr){4-5} \cmidrule(lr){6-6} \cmidrule(lr){7-8} \cmidrule(lr){9-9}
     & Acc@0.25 & Acc@0.5 & F1@0.25 & F1@0.5 & C@0.5 & C & EM & EM \\
    \midrule
    Baseline & 62.1 & 54.6 & 59.6 & 54.4 & 81.4 & 104.6 & 30.5 & 60.8 \\
    \midrule
    w/o Zero-init & 63.1 & 56.1 & 61.0 & 55.6 & 76.4 & 104.7 & 30.8 & 61.1 \\
    Vanilla Addition & 64.2 & 57.4 & 61.2 & 56.0 & 82.6 & 106.4 & 30.9 & 61.9 \\
    \midrule
    \textbf{ViPS (Full)} & \textbf{64.6} & \textbf{57.6} & \textbf{62.0} & \textbf{56.5} & \textbf{85.5} & \textbf{107.9} & \textbf{31.6} & \textbf{62.5} \\
    \bottomrule
    \end{tabular}
\end{table*}

%% file: tables/table5.tex
\begin{table*}[t]
    \centering
    \small
    \caption{\textbf{Ablation on the Efficient Prior Proxy.} The lightweight proxies match the upper bound (\colorbox{blue!15}{w/ GT Priors}, where features are extracted via independent forward passes of the original foundation models) with only a marginal performance drop, while reducing parameter and inference cost from $\sim$5$\times$ to 1$\times$.}
    \label{tab:ablation_epp}
    \setlength{\tabcolsep}{1.2pt}
    \begin{tabular}{l c c c c c c c c c c c}
    \toprule
    \multirow{2}{*}{\textbf{Method}} & \multicolumn{3}{c}{\textbf{Efficiency}} & \multicolumn{2}{c}{ScanRefer} & \multicolumn{2}{c}{Multi3DRefer} & S2C & \multicolumn{2}{c}{ScanQA} & SQA \\
    \cmidrule(lr){2-4} \cmidrule(lr){5-6} \cmidrule(lr){7-8} \cmidrule(lr){9-9} \cmidrule(lr){10-11} \cmidrule(lr){12-12}
     & \textbf{Param.} & \textbf{Latency} & \textbf{Err. $\downarrow$} & Acc@0.25 & Acc@0.5 & F1@0.25 & F1@0.5 & C@0.5 & C & EM & EM \\
    \midrule
    \cellcolor{blue!15}w/ GT Priors & \cellcolor{red!15}$\sim5\times$ & \cellcolor{red!15}$\sim5\times$ & 0 & \textbf{65.4} & \textbf{58.3} & \textbf{62.9} & \textbf{57.4} & \textbf{86.0} & \textbf{107.9} & \textbf{32.1} & \textbf{63.2} \\
    \midrule
     w/o $\mathcal{L}_{alignment}$ & \cellcolor{green!15}$1\times$ & \cellcolor{green!15}$1\times$ & - & 64.0 & 57.2 & 61.3 & 55.9 & 83.4 & 106.3 & 30.7 & 61.1 \\
     \textbf{ViPS (Ours)} & \cellcolor{green!15}$1\times$ & \cellcolor{green!15}$1\times$ & 0.252 & 64.6 & 57.6 & 62.0 & 56.5 & 85.5 & \textbf{107.9} & 31.6 & 62.5 \\
    \bottomrule
    \end{tabular}
\end{table*}

%% file: chapters/related_work.tex
\section{Related Work}
\subsection{Spatial Understanding with Large Language Models}
Spatial understanding, a foundational pillar for real-world interaction and reasoning, has witnessed a paradigm shift with the advent of LLMs~\citep{brown2020language, ouyang2022training, touvron2023llama,achiam2023gpt}. Early attempts, such as PointLLM~\citep{xu2024pointllm}, PointBind~\citep{guo2023point}, GPT4Point~\citep{qi2024gpt4point}, MiniGPT-3D~\citep{tang2024minigpt}, and Chat-3D~\citep{wang2023chat}, focused on aligning 3D point-cloud encoders directly with the LLM embedding space. To facilitate more effective cross-modal feature fusion, subsequent frameworks like Grounded-3D-LLM~\citep{chen2024grounded}, LL3DA~\citep{chen2024ll3da}, 3D-LLaVA~\citep{deng20253d}, and Inst3D-LLM~\citep{yu2025inst3d} introduced advanced representation learning schemes. However, the inherent scarcity and noise of 3D point-cloud data often limit the scalability of these methods.

Recent research has gravitated towards video-based inputs. Prominent works such as 3D-LLM~\citep{hong20233d}, Scene-LLM~\citep{fu2024scene}, Video-3D LLM~\citep{zheng2025video}, GPT4Scene~\citep{qi2025gpt4scene} and SpatialStack~\citep{zhang2026spatialstack} establish dense correlations between 2D features and 3D scenes by building upon powerful pre-trained MLLMs~\citep{li2024llava, wang2024qwen2}. Specifically, Scene-LLM~\citep{fu2024scene} captures fine-grained 3D knowledge through efficient 3D visual representation learning, while Video-3D LLM~\citep{zheng2025video} introduces position-aware encodings for video sequences. Similarly, LLaVA-3D~\citep{zhu2024llava} achieves robust perception by learning a set of 3D voxels. While our work also falls within the video-input MLLM paradigm, we diverge from these approaches by exploring how to synergistically integrate prior knowledge from diverse foundation models to further elevate 3D spatial understanding.
\subsection{Integration of Foundation Model Prior in MLLMs}
The rapid evolution of foundation models—such as VGGT~\citep{wang2025vggt}, DepthAnything3~\citep{lin2025depth}, and WAN~\citep{wan2025wan}—has inspired a new line of research that injects diverse foundational priors into MLLMs to bolster their perceptual capabilities. Recent methods including VG-LLM~\citep{zheng2025learning}, 3DRS~\citep{huang20253drs}, MiLO~\citep{cao2025seeing}, VLM-3R~\citep{fan2025vlm}, Vega-3D~\citep{wu2026generation}, ROSS3D~\citep{wang2025ross3d}, and GeoThinker~\citep{li2026thinking} have demonstrated the efficacy of this paradigm. Specifically, VG-LLM employs a 3D visual geometry encoder to extract geometric priors from video sequences, while 3DRS aligns the latent space of MLLMs with VGGT via knowledge distillation. GeoThinker introduces an active perception mechanism for MLLMs to retrieve necessary geometric features, and Vega-3D injects world knowledge from video generation models, predicated on the hypothesis that these models inherently capture the underlying dynamics of scene transitions. Besides MLLM-based approaches, CUA-O3D~\citep{li2025cross} uses uncertainty-aware distillation from multiple foundation models for open-vocabulary 3D semantic segmentation. In summary, existing MLLM-based approaches have predominantly focused on: (i) more efficient mechanisms for utilizing prior knowledge, and (ii) evaluating the performance of different individual model priors. In contrast, we propose \textbf{ViPS}, an efficient multi-model prior framework. By adaptively harmonizing various priors from disparate sources, our method significantly improves MLLM performance in spatial understanding tasks.

%% file: chapters/conclusion.tex
\section{Conclusion}

In this paper, we have introduced \textbf{ViPS}, a multi-model prior framework that unifies heterogeneous foundation-model priors within a single MLLM for spatial understanding. Our empirical study first reveals that no single foundation model dominates across tasks: different models supply distinct and complementary spatial knowledge. Building on this insight, ViPS couples an \emph{Efficient Prior Proxy}, which distills the knowledge of multiple foundation models into lightweight branches sharing a common backbone and thereby avoids the linear cost of independent forward passes, with a \emph{Dynamic Prior Fusion} mechanism that adaptively re-weights the resulting priors according to the input query and injects them through zero-initialized convolutions for stable training. On VSI-Bench and five ScanNet-series benchmarks, ViPS delivers state-of-the-art results in spatial reasoning, 3D visual grounding, dense captioning, and embodied question answering, while retaining single-encoder-level inference cost. These results suggest that context-aware harmonization of complementary priors is a promising path toward spatially grounded multimodal reasoning.

%% file: chapters/ack.tex
\section*{Acknowledgements}
This work is supported by Hong Kong Research Grants Council -- General Research Fund (Grant No.\ 17213825), Hong Kong Innovation and Technology Commission -- Innovation and Technology Fund (Grant No.\ ITS/488/24FP), and HKU Seed Fund for PI Research.

%% file: chapters/appendix.tex
\begin{center}
\LARGE \textbf{Appendix}
\end{center}
\vspace{0.5cm}

\section{Detailed Training Dataset Description}
\label{sec:training_datasets}

\subsection{ScanNet-series Dataset}
For the 3D spatial understanding tasks, our training protocol aligns with the configurations established by previous works such as Video-3D LLM and 3DRS, which aggregate data from five distinct benchmarks, yielding a combined training set of approximately 223K sample pairs. Specifically, the composition includes:
\begin{itemize}
    \item \textbf{ScanRefer and Scan2Cap:} Each dataset contributes 36,665 QA instances.
    \item \textbf{Multi3DRefer:} This multi-target grounding dataset provides 43,838 data entries.
    \item \textbf{ScanQA:} We utilize 26,515 question-answering pairs from this dataset.
    \item \textbf{SQA3D:} Serving as the largest subset in our collection, SQA3D contributes 79,445 samples.
\end{itemize}
With the exception of SQA3D, which is sourced from 518 unique 3D environments, the remaining four datasets are constructed upon 562 unique scans from the ScanNet corpus. 

\subsection{VSI-Bench Dataset}
For the spatial reasoning evaluation, we follow the dataset composition presented in VLM-3R to train our model on VSI-Bench. Our training subset comprises a total of 207,658 instruction-tuning QA pairs. The dataset is structurally diverse, encompassing various scene typologies and specific reasoning tasks. The detailed distribution is as follows:
\begin{itemize}
    \item \textbf{ScanNet++ QA:} The majority of the data originates from the high-fidelity ScanNet++ corpus, accounting for 135,119 QA pairs.
    \item \textbf{ScanNet QA:} An additional 51,630 QA pairs are derived from ScanNet.
    \item \textbf{Absolute Distance Estimation:} 16,805 samples are dedicated to object absolute distance estimation tasks.
    \item \textbf{Route Planning:} The dataset also includes 4,104 instances specifically designed to evaluate and enhance the model's route planning and navigational reasoning capabilities.
\end{itemize}

\subsection{Details of Section~\ref{sec:motivation}}
\label{sec:motivation_details}
To ensure computational efficiency during our exploration, here we train the models on a 10\% subset of the original training datasets used in 3DRS~\citep{huang20253drs} and VLM-3R~\citep{fan2025vlm}. We conduct our empirical assessment on the VSI-Bench~\citep{yang2025thinking} and ScanNet-series benchmarks including ScanRefer~\citep{chen2020scanrefer}, Multi3DRefer~\citep{zhang2023multi3drefer}, Scan2Cap~\citep{chen2021scan2cap}, ScanQA~\citep{azuma2022scanqa}, and SQA3D~\citep{ma2022sqa3d}. We employ Qwen2-VL~\citep{wang2024qwen2} as our base MLLM. For each evaluated variant, we extract features from each foundation model via its respective encoder and project them into the same hidden dimension as the MLLM using an MLP. Then inject them into the image tokens of the MLLM via addition. The evaluation metrics are consistent with those in the main text.

\section{Training Detail}
\label{sec:training_details}

A comprehensive summary of all detailed hyperparameters and configurations is provided in Table~\ref{tab:hyperparameters_qwen2} and Table~\ref{tab:hyperparameters_qwen3}. To align the external visual priors with the internal representations, we downsample all features extracted by the prior models to match the resolution of the MLLM's image tokens.

The trainable parameters in our ViPS framework include the LLM backbone (updated via LoRA), the multimodal projector (\texttt{mm\_projector}), the zero-initialized convolutions and MLPs responsible for generating dynamic weights within the Dynamic Prior Fusion, and all parameters within the Efficient Prior Proxy. Under our hardware configuration, the training process takes approximately 24 hours for the ScanNet-series datasets and 30 hours for the VSI-Bench dataset.

\input{tables/table_6.tex}
\input{tables/table_7.tex}

\clearpage
\section{Qualitative Analysis of Visual Priors}
\label{sec:prior_analysis}

\begin{figure}[htbp]
  \centering
  \includegraphics[width=\linewidth]{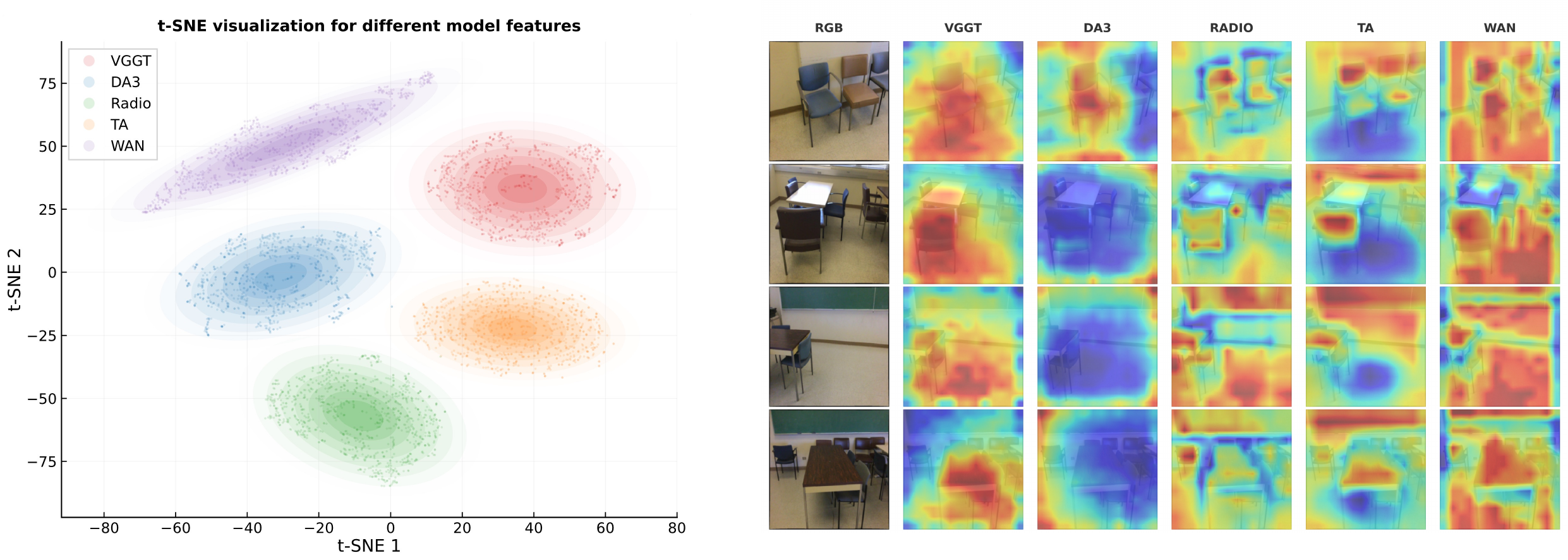}
  \caption{\textbf{Prior Analysis of Diverse Foundation Models.} \textbf{Left:} The t-SNE of visual prior features extracted from different foundation models. \textbf{Right:} Spatial heatmaps for different models}
  \label{fig:feature_ana}
\end{figure}

We conduct a prior analysis on the diverse foundation models evaluated in Section~\ref{sec:motivation}. As illustrated in Figure~\ref{fig:feature_ana} (Left), the visualization demonstrates that priors extracted from distinct foundation models cluster into disparate regions within the latent space. Furthermore, the activation heatmaps in Figure~\ref{fig:feature_ana} (Right) reveal that different models focus on varying structural and semantic cues within the identical scene. These results validate that different foundation models indeed encapsulate distinct representations and spatial cues due to their diverse training paradigms.

\section{Impact of Different Base Model}
\label{sec:base_models}

In the main paper, we default to using VGGT as the base foundation model within our Efficient Prior Proxy. To further investigate the robustness and flexibility of our ViPS framework, we conduct additional experiments by substituting VGGT with other foundation models as the base model, while keeping the rest of the framework unchanged.

As shown in Table~\ref{tab:base_models}, using different base models yields comparable and consistently strong performance across the ScanNet-series benchmarks. Although VGGT yields the best overall performance, we observe that TraceAnything achieves better results on certain metrics (e.g., 62.02 F1@0.25 on Multi3DRefer and 85.53 C@0.5 on Scan2Cap). Furthermore, all other foundation models maintain highly comparable performance when used as the base model. These results demonstrate that our Efficient Prior Proxy and Dynamic Prior Fusion mechanisms are robust and not strictly dependent on a specific base model choice.

\input{tables/table_base_models.tex}

\section{Visualization Result}
\label{sec:visualization}

\begin{figure}[htbp]
  \centering
  \includegraphics[width=0.9\linewidth,height=0.9\textheight,keepaspectratio]{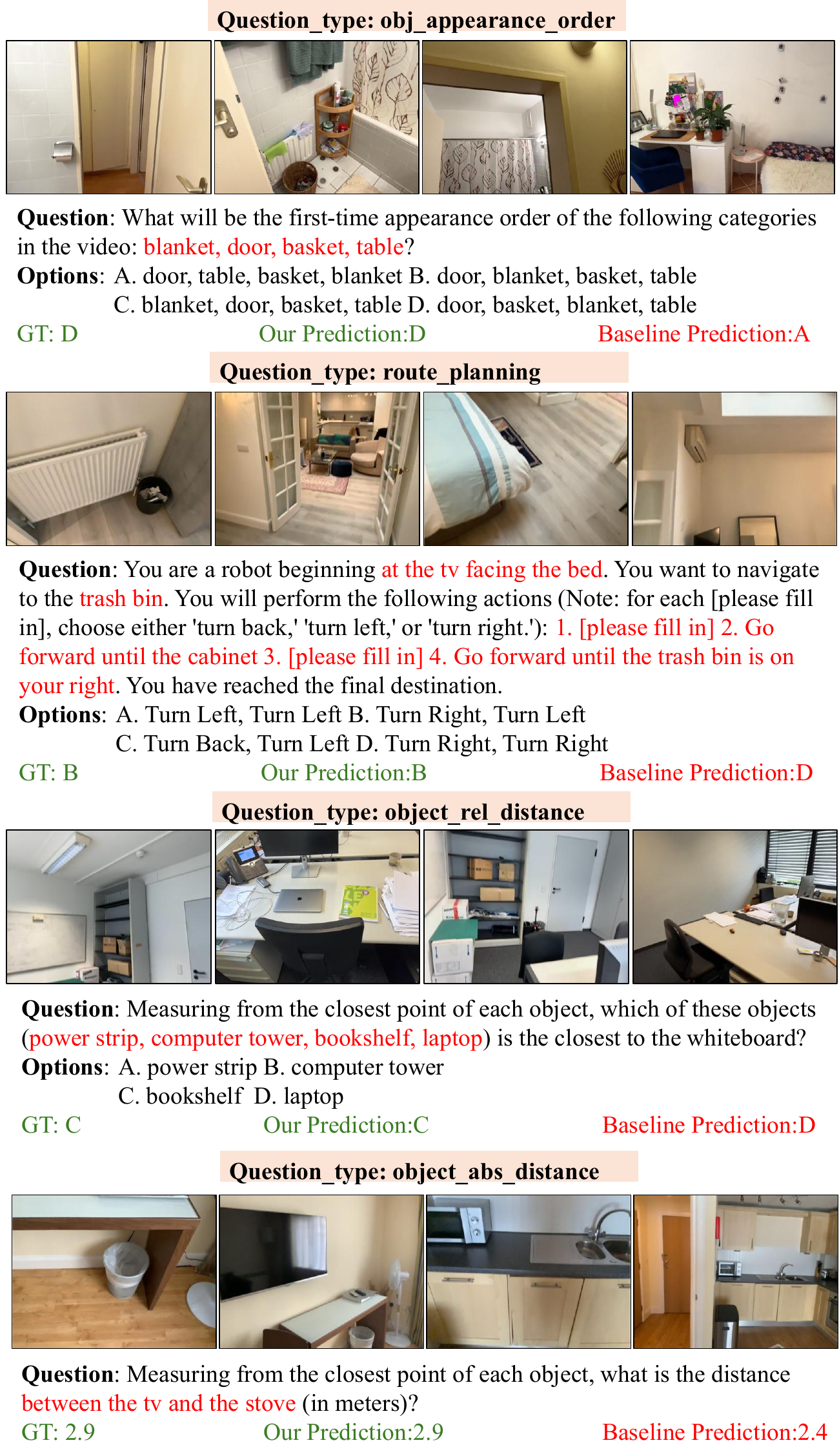}
  \caption{\textbf{Qualitative Visualization on VSI-Bench.} We compare our ViPS framework against the baseline, which is obtained by fine-tuning \texttt{Qwen2-VL} on our identical training dataset, showing that ViPS produces more accurate responses to complex spatial questions.}
  \label{fig:vsibench_viz}
\end{figure}

To further demonstrate the effectiveness of our proposed ViPS framework, we provide visualization results in Figure~\ref{fig:vsibench_viz}. In this comparison, the baseline model is obtained by fine-tuning the \texttt{Qwen2-VL} architecture on the same training dataset used for our model. As illustrated, by synergistically harmonizing diverse visual priors, our ViPS framework achieves a more precise understanding of complex spatial relationships and generates more accurate responses compared to the baseline.

\section{Broader Impact}
\label{sec:broader_impacts}

\textbf{Positive Impacts.} The capabilities introduced by our ViPS framework significantly advance the spatial reasoning and 3D spatial understanding of MLLMs. By effectively harmonizing diverse visual priors, our approach paves the way for more intelligent embodied agents, such as domestic robots and autonomous navigation systems. Furthermore, our Efficient Prior Proxy minimizes inference overhead, facilitating the deployment of sophisticated 3D perception systems on resource-constrained edge devices. This efficiency can democratize access to advanced AI, fostering innovations in assistive applications for visually impaired individuals to navigate complex environments safely.

\textbf{Potential Negative Impacts and Mitigation.} Endowing AI systems with highly accurate 3D spatial understanding capabilities inherently presents privacy and security risks. If deployed irresponsibly, such technologies could be exploited for unauthorized surveillance or the invasive reconstruction of private indoor spaces. To mitigate these ethical concerns, future real-world applications must incorporate robust anonymization pipelines—such as the automatic obfuscation of human faces and sensitive documents—prior to processing scene data. Additionally, strict access controls, transparent user consent mechanisms, and compliance with data protection regulations are essential to safeguard individual privacy.

\section{Limitation and Future Work}
\label{sec:limitations}

While ViPS demonstrates state-of-the-art performance in 3D spatial understanding, several limitations remain. First, due to computational resource constraints, our experiments are currently restricted to a 7B-parameter base MLLM and training datasets on the scale of a few hundred thousand samples. Scaling up both the model size and data volume could further enhance performance. Second, extracting spatial awareness from external visual priors is ultimately a sub-optimal workaround. A more fundamental future direction is to internalize this geometric knowledge directly during the MLLM pre-training phase, which could be achieved by introducing more diverse spatial datasets, fine-grained 3D sub-tasks, and tailored loss functions. Finally, our evaluations are currently confined to standard offline benchmarks. Deploying and validating spatial-aware MLLMs in real-world robotics and physical embodied AI scenarios remains an exciting avenue for future exploration.

%% file: tables/table_6.tex
\begin{table}[h]
\centering
\caption{\textbf{Training Hyperparameters and Hardware Configuration.} Detailed settings used to train ViPS (qwen2-vl-7b) on the ScanNet-series benchmarks.}
\label{tab:hyperparameters_qwen2}
\resizebox{0.8\textwidth}{!}{%
\begin{tabular}{lc}
\toprule
\textbf{Configuration / Hyperparameter} & \textbf{Value} \\
\midrule
\multicolumn{2}{l}{\textit{Model Initialization}} \\
Base MLLM & \texttt{LLaVA-Video-7B-Qwen2} \\
Vision Tower & \texttt{google/siglip-so400m-patch14-384} \\
\midrule
\multicolumn{2}{l}{\textit{Training Settings}} \\
Hardware & 8 $\times$ 48GB GPUs \\
Distributed Strategy & DeepSpeed ZeRO Stage 2 \\
Data Precision & \texttt{bfloat16} (bf16) \\
Gradient Checkpointing & True \\
\midrule
\multicolumn{2}{l}{\textit{Optimization Hyperparameters}} \\
Training Epochs & 1 \\
Per-device Batch Size & 1 \\
Gradient Accumulation Steps & 2 \\
Total Effective Batch Size & 16 \\
Base Learning Rate & $2 \times 10^{-5}$ \\
Warmup Ratio & 0.03 \\
\midrule
\multicolumn{2}{l}{\textit{LoRA Configurations}} \\
LoRA Rank & 512 \\
LoRA Alpha & 1024 \\
\midrule
\multicolumn{2}{l}{\textit{Prior Features Alignment}} \\
Target Spatial Resolution & $14 \times 14$ \\
Downsampling Method & Bilinear \\
\bottomrule
\end{tabular}%
}
\end{table}

%% file: tables/table_7.tex
\begin{table}[h]
\centering
\caption{\textbf{Training Hyperparameters and Hardware Configuration.} Detailed settings used to train ViPS (qwen3-vl-8b) on the ScanNet-series and VSI-Bench datasets.}
\label{tab:hyperparameters_qwen3}
\resizebox{0.8\textwidth}{!}{%
\begin{tabular}{lc}
\toprule
\textbf{Configuration / Hyperparameter} & \textbf{Value} \\
\midrule
\multicolumn{2}{l}{\textit{Model Initialization}} \\
Base MLLM & \texttt{Qwen/Qwen3-VL-8B-Instruct} \\
Vision Tower & \texttt{Qwen3VLVisionModel} \\
\midrule
\multicolumn{2}{l}{\textit{Training Settings}} \\
Hardware & 8 $\times$ 48GB GPUs \\
Distributed Strategy & DeepSpeed ZeRO Stage 2 \\
Data Precision & \texttt{bfloat16} (bf16) \\
Gradient Checkpointing & True \\
\midrule
\multicolumn{2}{l}{\textit{Optimization Hyperparameters}} \\
Training Epochs & 1 \\
Per-device Batch Size & 1 \\
Gradient Accumulation Steps & 2 \\
Total Effective Batch Size & 16 \\
Base Learning Rate & $2 \times 10^{-5}$ \\
Warmup Ratio & 0.03 \\
\midrule
\multicolumn{2}{l}{\textit{LoRA Configurations}} \\
LoRA Rank & 256 \\
LoRA Alpha & 256 \\
\midrule
\multicolumn{2}{l}{\textit{Prior Features Alignment}} \\
Target Spatial Resolution & $15 \times 20$ \\
Downsampling Method & Bilinear \\
\bottomrule
\end{tabular}%
}
\end{table}

%% file: tables/table_base_models.tex
\begin{table}[h]
\centering
\caption{\textbf{Robustness to the Choice of Base Model.} Substituting VGGT with other foundation models in the Efficient Prior Proxy yields comparable performance across the ScanNet-series benchmarks, indicating that ViPS is not strictly dependent on a specific base-model choice.}
\label{tab:base_models}
\resizebox{\textwidth}{!}{%
\begin{tabular}{l c c c c c c c c}
\toprule
Base Model & \multicolumn{2}{c}{ScanRefer} & \multicolumn{2}{c}{Multi3DRefer} & \multicolumn{1}{c}{Scan2Cap} & \multicolumn{2}{c}{ScanQA} & \multicolumn{1}{c}{SQA3D} \\
\cmidrule(lr){2-3} \cmidrule(lr){4-5} \cmidrule(lr){6-6} \cmidrule(lr){7-8} \cmidrule(lr){9-9}
& Acc@0.25 & Acc@0.5 & F1@0.25 & F1@0.5 & C@0.5 & CIDER & EM & EM \\
\midrule
Baseline & 62.1 & 54.6 & 59.6 & 54.4 & 81.4 & 104.6 & 30.5 & 60.8 \\
\midrule
VGGT & \textbf{64.6} & \textbf{57.6} & 62.0 & \textbf{56.5} & 85.5 & \textbf{107.9} & \textbf{31.6} & \textbf{62.5} \\
Wan & 64.4 & 57.4 & 62.0 & 56.3 & 83.0 & 106.0 & 31.1 & 61.4 \\
TraceAnything & 63.8 & 56.7 & \textbf{62.0} & 56.4 & \textbf{85.5} & 107.0 & 31.3 & 61.8 \\
DepthAnything & 63.7 & 56.8 & 61.8 & 56.4 & 82.2 & 107.0 & 31.1 & 61.7 \\
RADIO & 64.1 & 57.2 & 61.7 & 56.4 & 82.6 & 107.3 & 31.2 & 62.5 \\
\bottomrule
\end{tabular}%
}
\end{table} 